\pdfoutput=1

\documentclass{article}
\usepackage{ijcai24}

\usepackage{times}
\usepackage{soul}
\usepackage{url}
\usepackage[hidelinks]{hyperref}
\usepackage[utf8]{inputenc}
\usepackage[small]{caption}
\usepackage{graphicx}
\usepackage{amsmath}
\usepackage{amsthm}
\usepackage{booktabs}
\usepackage{algorithm}
\usepackage{algorithmic}
\usepackage[switch]{lineno}
\usepackage{amsfonts}

\usepackage{multirow}

\title{Decoupled and Interactive Regression Modeling for High-performance One-stage 3D Object Detection}

\author{
    Author Name
    \affiliations
    Affiliation
    \emails
    email@example.com
}

\author{
Weiping Xiao$^{1,2\ast}$
\and
Yiqiang Wu$^{1,2\ast}$\and
Chang Liu$^{{1,2}}$\And
Yu Qin$^{1,2}$\And
Xiaomao Li$^{1,2}$\And
Liming Xin$^1$\\
\affiliations
$^1$Shanghai University, $^2$Engineering Research Center of Unmanned Intelligent Marine Equipment of the Ministry of Education\\
\emails
\{weiping.xiao, yiqiangwu, liuchang123, qy21722148\}@shu.edu.cn,
{lixiaomaosia}@163.com,
{limingxin}@shu.edu.cn
}

\begin{document}

\maketitle

\begin{abstract}
Inadequate bounding box modeling in regression tasks constrains the performance of one-stage 3D object detection. Our study reveals that the primary reason lies in two aspects: (1) The limited center-offset prediction seriously impairs the bounding box localization since many highest response positions significantly deviate from object centers. (2) The low-quality sample ignored in regression tasks significantly impacts the bounding box prediction since it produces unreliable quality (IoU) rectification. To tackle these problems, we propose Decoupled and Interactive Regression Modeling (DIRM) for one-stage detection. Specifically, Decoupled Attribute Regression (DAR) is implemented to facilitate long regression range modeling for the center attribute through an adaptive multi-sample assignment strategy that deeply decouples bounding box attributes. On the other hand, to enhance the reliability of IoU predictions for low-quality results, Interactive Quality Prediction (IQP) integrates the classification task, proficient in modeling negative samples, with quality prediction for joint optimization. Extensive experiments on Waymo and ONCE datasets demonstrate that DIRM significantly improves the performance of several state-of-the-art methods with minimal additional inference latency. Notably, DIRM achieves state-of-the-art detection performance on both the Waymo and ONCE datasets.
\end{abstract}

\renewcommand{\thefootnote}{}
\footnotetext{$\ast$Equal contribution}

\section{Introduction}
\label{sec:intro}
With the widespread application of LiDAR in autonomous driving, LiDAR-based 3D object detection garners increasing attention and substantial development. 
Current high-performance 3D detectors commonly adopt two-stage network structures. 
In comparison to one-stage competitors, two-stage methods involve additional time-consuming operations, such as Set Abstraction \cite{pointnet2} and Region of Interest (RoI) Pooling \cite{fastrcnn}, and impose higher memory burdens, restricting their applicability in real-world autonomous driving scenes.

\begin{figure}[t]
\centering
\includegraphics[width=0.9\columnwidth]{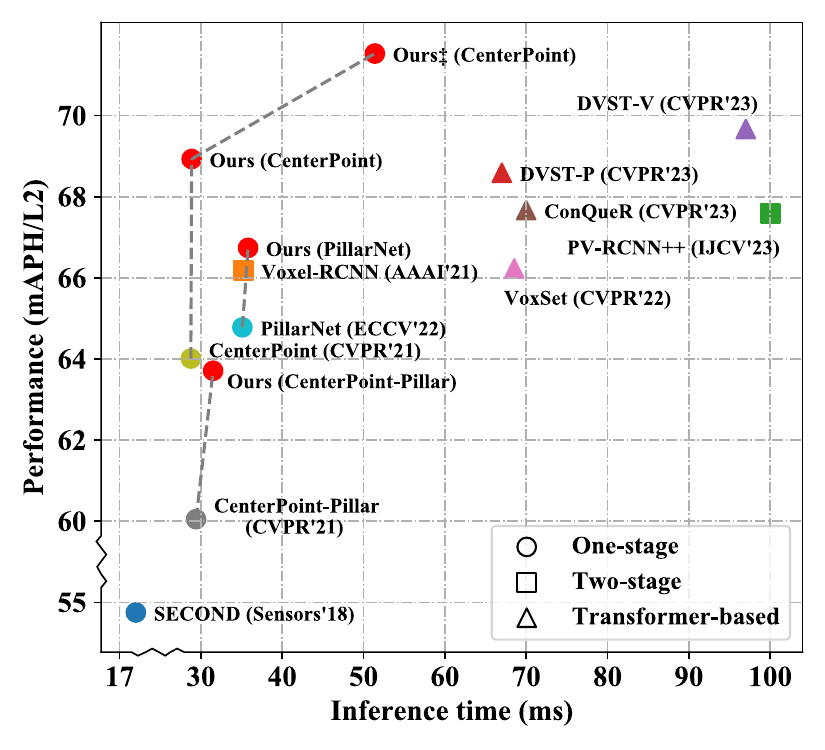}
\caption{Detection performance and inference time of DIRM, compared with state-of-the-art methods. All methods are trained on the 20\% Waymo training set, and the inference latency is evaluated on a single NVIDIA A100 GPU. Results show that DIRM remarkably outperforms the baseline method while adding little inference latency. Besides, DIRM outperforms the previous state-of-the-art two-stage and transformer-based methods.}
\label{fig1}
\end{figure}

Existing one-stage methods strive to narrow the performance gap with two-stage ones. 
For instance, CIA-SSD \cite{ciassd} attempts to address the ambiguity between confidence and localization quality by introducing the Intersection over the Union (IoU) branch. 
PillarNet \cite{pillarnet} adopts a deeper encoder network and orientation decoupled DIoU loss, further optimizing detection capability. 
Although achieving decent accuracy gains, these methods still fall short of state-of-the-art (SOTA) two-stage methods. 
Recently, several methods \cite{dsvt,sst,voxelset} introduce self-attention \cite{transformer} and cross-attention into current dense 3D object detection. 
Despite these transformer-based methods deliver satisfactory performance, they incur expensive computational costs. 
Thus, designing a real-time and high-performance one-stage detector remains a challenging task.

To fully exploit the performance potential of one-stage methods, we conduct comprehensive analyses and experiments on the prevailing CenterPoint \cite{centerpoint} and reveal that inadequate modeling of regression tasks is the primary reason for the suboptimal performance.

\begin{figure}[t]
\centering
\includegraphics[width=\columnwidth]{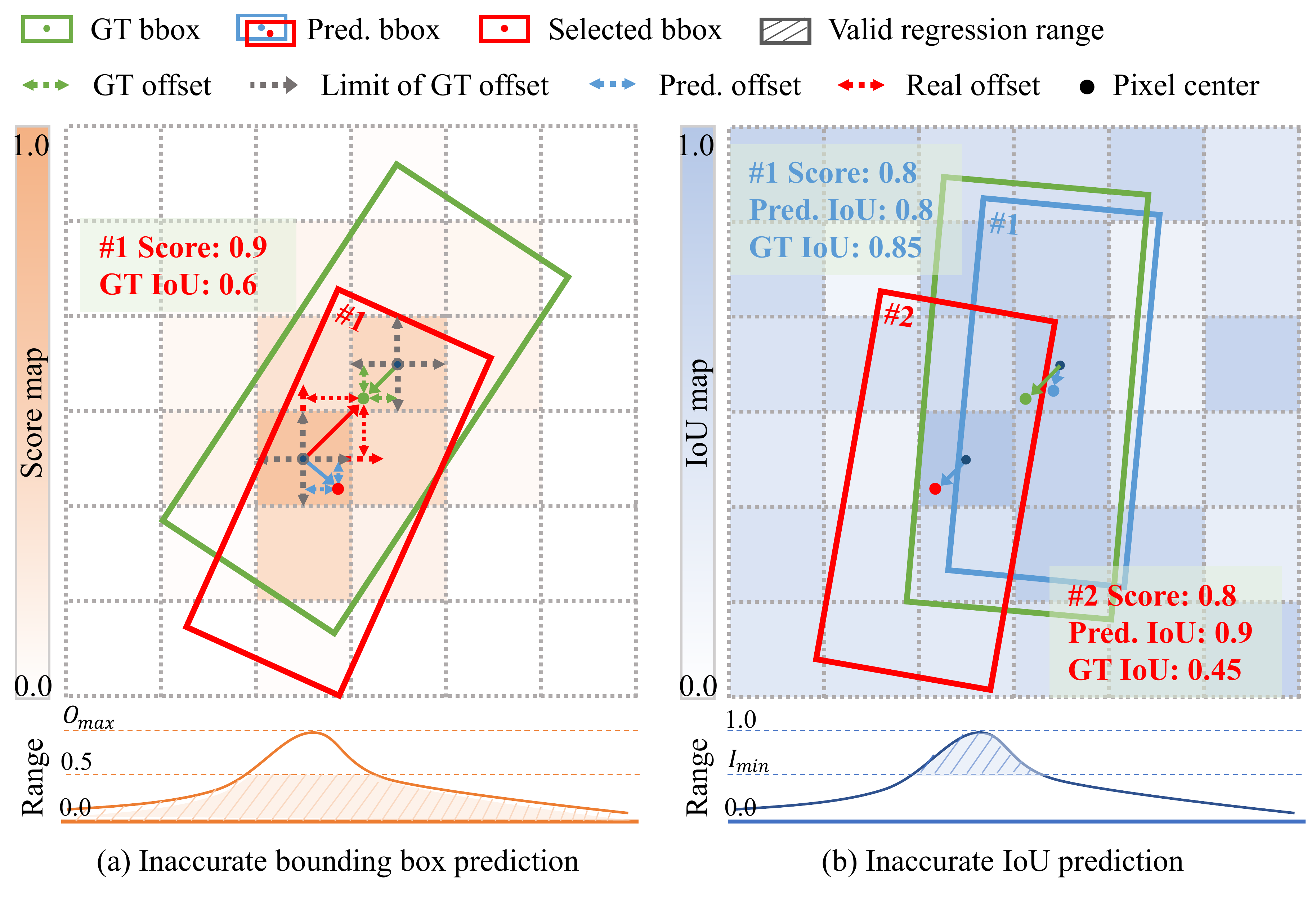}
\caption{Schematic representation of incorrect center attribute regression and IoU prediction. $O_{max}$ represents the greatest offset error of the center attribute, and $I_{min}$ is the minimum value of the IoU prediction.}
\label{fig2}
\end{figure}

{\em \textbf{Inaccurate center attribute regression}.} Center attribute regression as the core task of bounding box regression, focuses on the offset between the pixel center and the ground truth (GT) center. As shown in Fig. \ref{fig2} (a), the regression of the center attribute is modeled within a narrow interval (±0.5 pixels). Due to the insufficient modeling of long-range regression intervals, when the highest response deviates far from the object center, the limited prediction offset can significantly impact the localization of the bounding box. According to our statistics, the proportion of this phenomenon is nearly 70\%, and the mean relative percentage error (MRPE) of the center attribute reaches 130\%. 

{\em \textbf{Inaccurate quality (IoU) prediction}.} As another critical regression task, the quality of the bounding box (IoU) is predicted to rectify confidence scores. Similar to other regression tasks, IoU prediction focuses on the central samples of GTs, lacking reliable modeling for surrounding low-quality samples. Unreliable predicted IoUs can disrupt the rectification process as shown in Fig. \ref{fig2} (b). Further statistical results show that the mean square error (MSE) of the predicted IoU values for low-quality bounding boxes ($IoU_{GT}\leq 0.5$) is 22 times that for high-quality bounding boxes ($IoU_{GT} \textgreater 0.5$).

\begin{figure*}[t]
\centering
\includegraphics[width=0.8\textwidth]{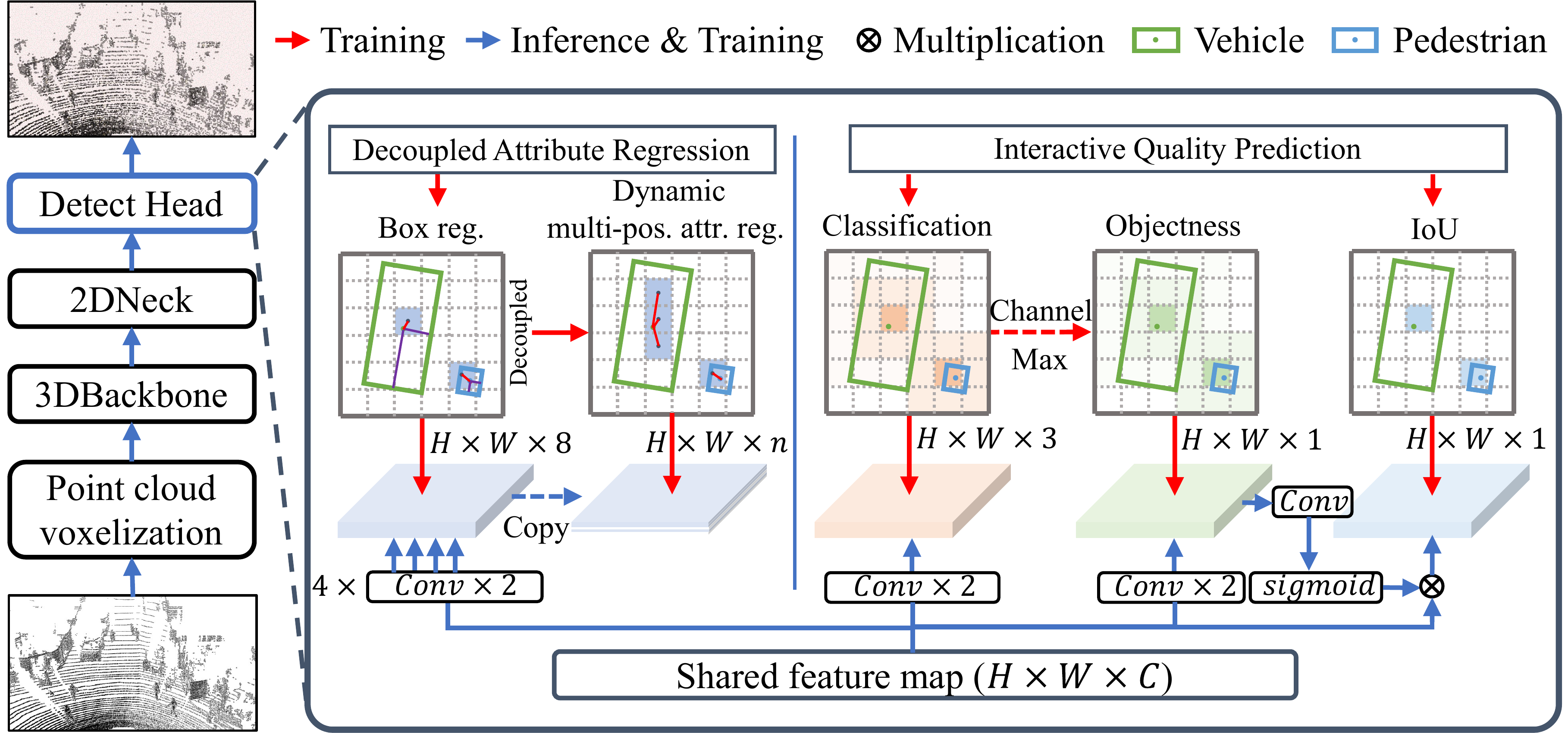}
\caption{The DIRM framework. The network contains two key designs, namely Decoupled Attribute Regression (DAR) and Interactive Quality Prediction (IQP). In the training stage, DAR only performs adaptive multi-sample assignment and supervision on several bounding box attributes, while IQP performs joint optimization and supervision on the class-agnostic object classification and quality prediction tasks. $Conv$ represents a conventional 2D convolutional layer, where the size of the convolution kernel is 3 × 3, and the number of channels remains the same as that of the shared feature map.}
\vspace{-0.35cm} 
\label{fig3}
\end{figure*}

To address these issues, we propose Decoupled and Interactive Regression Modeling (DIRM) for accurate bounding box regression and quality (IoU) prediction (Fig. \ref{fig3}). Specifically, for the center attribute, the Decoupled Attribute Regression (DAR) strategy is proposed to perform long-range regression modeling with the center and surrounding samples. Different from conventional multi-positive sample strategies, DAR deeply decouples the center attribute from other bounding box attributes and implements a parallel sample selection strategy as shown in Fig. \ref{fig4} (a). Thus, DAR can perform targeted modeling on some attributes and effectively avoid performance skew caused by imbalanced samples. For sample selection, DAR first picks initial samples according to point cloud distribution (position and orientation). Dynamic adjustment and optimization are then conducted based on the performance of these samples to collect solid regression clues in different training stages. On the other hand, to provide dependable predicted IoUs for low-quality predictions, the Interactive Quality Prediction (IQP) strategy is proposed to cleverly introduce a class-agnostic object classification into the IoU prediction task. Owing to the dense binary supervision signal, IQP can comprehensively model the foreground and background samples of the object, handling the disadvantages of past methods that are unable to supervise background samples. Based on this, IQP further finely optimizes the object foreground with sparse-quality supervision signals. With this interactive modeling, IQP can meet IoU prediction requirements of different quality results.

The proposed DAR and IQP can easily integrate into any center-based method with little additional complexity. Extensive experiments on the Waymo \cite{waymo} and ONCE \cite{once} datasets demonstrate that DIRM can significantly enhance existing one-stage SOTA methods by 2.0$\sim$5.0 mAPH and achieve new SOTA performance.

The main contributions of this work can be summarized as follows:

\begin{itemize}
    \item We introduce DIRM, a one-stage detector that achieves real-time and high-performance results by employing a decoupled and interactive regression modeling strategy for accurate bounding box regression and IoU prediction.
    \item DAR effectively models a long regression range for the center attribute by deeply decoupling bounding box attributes and implementing independent adaptive sample assignment strategies. To ensure dependable IoU predictions for low-quality results, IQP interactively models the class-agnostic object classification task and the quality prediction task.
    \item Extensive experiments conducted on both Waymo and ONCE datasets showcase the SOTA detection performance and superior generalization capabilities of DIRM. Additionally, quantitative experiments validate the effectiveness of DIRM in handling incomplete regression task modeling.
\end{itemize}

\section{Related Works}

Owing to more simplified networks, one-stage 3D object detection has garnered increased attention in academia and industry. However, one-stage methods consistently exhibit inferior performance compared to two-stage competitors. To enhance the performance of one-stage pipelines, the AFDet series \cite{afdetv2} introduces a quality prediction branch to rectify confidence scores, PillarNet \cite{pillarnet} increases the depth of the backbone network and introduces DIoU loss. PillarNext \cite{pillarnext} optimizes the performance of one-stage methods by redesigning the backbone, neck, and detection head. Recent methods \cite{voxelset,conquer,dsvt} integrate transformers \cite{transformer} into dense 3D object detection tasks, achieving advanced performance but incurring high computational costs.

{\em \textbf{Boundary box regression}.} Due to the outstanding performance of CenterPoint \cite{centerpoint} on large-scale datasets \cite{waymo,once}, various anchor-free methods adopt it as a new baseline \cite{afdet,afdetv2,pvrcnn++,pillarnet,centerformer,pillarnext,dsvt}. As a center-based detection method, CenterPoint constructs a regression model of bounding box attributes based on the central sample. As analyzed in Sec. \ref{sec:intro}, the regression model, built on center samples, struggles to provide precise localization, limiting the performance of center-based one-stage detection. This work conducts an in-depth analysis of inaccurate localization in one-stage center-based detection. The proposed DAR models the long-range regression for the center attribute, ensuring accurate localization for high-quality prediction results that are off-center.

{\em \textbf{Quality prediction}.} Several center-based methods \cite{pillarnet,afdetv2,dsvt,pillarnext} rectify the confidence scores by introducing an additional quality (IoU) prediction branch, aiming to retain predictions that are highly correlated with localization quality during the NMS process. These methods only focus on the central sample of the GT box, lacking reliable modeling of surrounding low-quality samples. Unreliable quality prediction results can easily interfere with the confidence correction process and directly impact the localization performance of the detector. The proposed IQP aims to integrate classification tasks proficient in modeling background information with regression tasks to improve the IoU prediction accuracy of different quality samples through joint optimization of multiple loss functions.

\section{DIRM}

\subsection{Decoupled Attribute Regression (DAR)}

Localization accuracy is a critical factor that impacts the quality of bounding box regression. Constrained by the unified center-based sample assignment, the regression target $d_{xy}$ of the center point is modeled within a limited pixel interval ($|d_{xy}| \le 0.5$). As discussed earlier, incomplete modeling for the center attribute results in significant localization deviation. Statistical results reveal that the relative error of the center attribute is 130\% for all categories and even 160\% for the vehicle category.

To model longer-range regression, a natural idea is to use multiple samples around the center point to predict the object's center, a strategy known as the multi-positive sample assignment. In this case, the regression target can be modeled within the interval of ($|d_{xy}| \textgreater 0.5$). However, simple experiments demonstrate that selecting samples around the center point as positive samples cannot lead to an overall performance improvement. This is because adding additional regression tasks to some attributes that are not sensitive to multi-sample will result in an imbalanced regression loss. To address this limitation, we propose a decoupled attribute regression (DAR) strategy, which includes the following crucial designs:

{\em \textbf{Deep Decoupling of Attributes}}. Although the center-based method divides the bounding box regression into different attribute regression tasks, all these attributes still adopt a unified sample assignment strategy and loss calculation as:
\begin{equation}
\label{equation_1}
L_{CT}=L1(F_{center\_pos}(\{R_{center},R_z,R_{lwh},R_{\theta}\})),
\end{equation}
where $\{R_{center},R_z,R_{lwh},R_{\theta}\}$ denotes four regression tasks for bounding box attributes, $F_{center\_pos}$ is the center-based sample assignment strategy, $L1$ is the L1 loss function, and $L_{CT}$ is the regression loss of center-based methods.

\begin{figure}[t]
    \centering
    \includegraphics[width=\columnwidth]{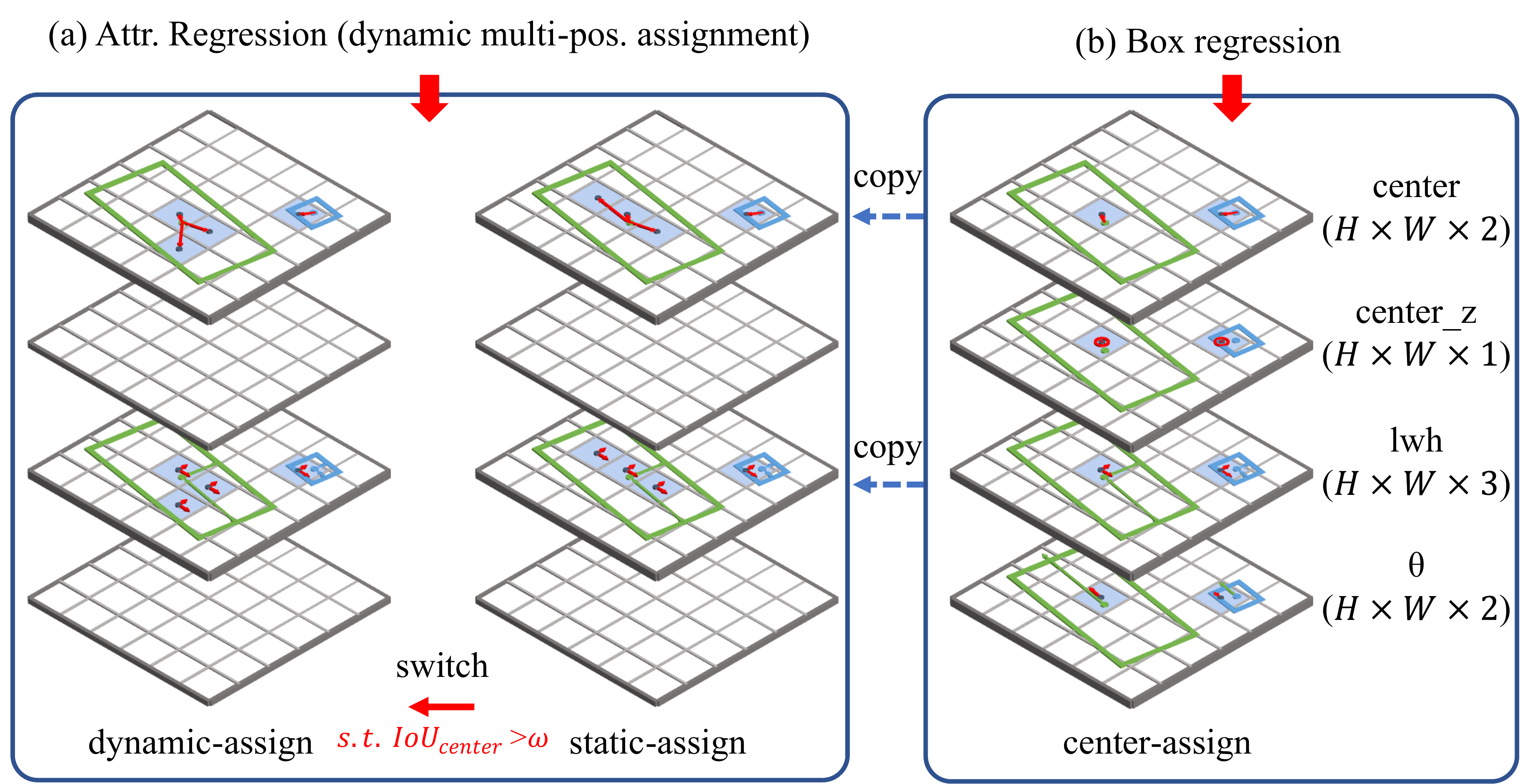}
    \caption{Decoupled attribute regression. "center", "center z", "lwh", and "$\theta$" respectively denote the center, center height, length, width, height, and orientation angle of the bounding box. $IoU_{center}$ is a good representation of the center sample's quality.}
    \vspace{-0.25cm} 
    \label{fig4}
\end{figure}

To overcome this limitation, DAR deeply decouples the bounding box attributes. It can be flexibly applied to different attribute combinations and build independent sample assignment strategies for them. Given a real bounding box $T_{box_i}$ including 7 attributes of $\{x,y,z,l,w,h,\theta\}$, as shown in Fig. \ref{fig4}. DAR can be applied only to $x$ and $y$ (2D center of the object), or it can be freely combined with other attributes. The regression loss of DAR is calculated as follows:
\begin{equation}
\label{equation_2}
L_{DAR}=L1(F_{DAR}(\{R_x,R_y,R_{attr}|attr \in \{z,l,w,h,\theta\}\})).
\end{equation}

This approach allows DAR to selectively model regression on specific attributes while preventing unnecessary redundancy in regression tasks.

{\em \textbf{Dynamic Sample Selection}}. Determining the appropriate samples for modeling with long regression ranges becomes the subsequent task. It is acknowledged that object points are sparse and non-uniform. Among the numerous candidate samples surrounding the center point of objects, only a small subset contains abundant points. Based on this observation, DAR selects initial samples primarily according to the distribution characteristics of the point cloud in the early training stages (static assign). DAR picks samples with richer point clouds based on the orientation of the object and its position relative to the LiDAR sensor. Please refer to the supplemental for more design details.

During training, DAR evaluates the performance of the center sample using IoU. Once the network achieves stable prediction ability, i.e., the predicted IoU performance of the center sample beyond threshold $IoU_{th}$, DAR will select top $k$ optimal samples based on the dynamic IoU performance of the samples around the center point (Fig. \ref{fig4}). The number of samples selected in the stable period remains the same as that in the initial period. In this way, DAR can adequately capture samples with rich regression cues for efficient modeling with long-range regression intervals.

With the above design, the long-range regression model established by DAR can provide more accurate localization for the bounding box regression task. Importantly, DAR introduces little inference time and memory consumption.

\begin{table*}[t]
	\centering
	\resizebox{0.85\linewidth}{!}{
		\begin{tabular}{l l c cc cc cc cc}
			\hline
			\multirow{2}{*}{Methods} & \multirow{2}{*}{Stage} & \multirow{2}{*}{FPS} & \multicolumn{2}{c}{mAP/mAPH} & \multicolumn{2}{c}{Vehicle AP/APH} &  \multicolumn{2}{c}{Pedestrian} &  \multicolumn{2}{c}{Cyclist AP/APH} \\
			& & & L1 & L2 & L1 & L2 & L1 & L2 & L1 & L2 \\
			\hline
			SECOND\cite{second} & One       & -      & 67.2/63.1 & 61.0/57.2 & 72.3/71.7 & 63.9/63.3 & 68.7/58.2 & 60.7/51.3 & 60.6/59.3 & 58.3/57.0 \\
			PointPillars\cite{pointpillars} & One        & -      & 69.0/63.5 & 62.8/57.8 & 72.1/71.5 & 63.6/63.1 & 70.6/56.7 & 62.8/50.3 & 64.4/62.3 & 61.9/59.9 \\
			RangeDet\cite{rangedet} & One        & -      & 71.5/69.5 & 65.0/63.2 & 72.9/72.3 & 64.0/63.6 & 75.9/71.9 & 67.6/63.9 & 65.7/64.4 & 63.3/62.1 \\
			CenterPoint\cite{centerpoint} & One        & 34.8  & 74.4/71.7 & 68.2/65.8 & 74.2/73.6 & 66.2/65.7 & 76.6/70.5 & 68.8/63.2 & 72.3/71.1 & 69.7/68.5 \\
			VoxSet\cite{voxelset} & One        & -      & 75.4/72.2 & 69.1/66.2 & 74.5/74.0 & 66.0/65.6 & 80.0/72.4 & 72.5/65.4 & 71.6/70.3 & 69.0/67.7 \\
			SST\cite{sst}   & One        & -      & 74.5/71.0 & 67.8/64.6 & 74.2/73.8 & 65.5/65.1 & 78.7/69.6 & 70.0/61.7 & 70.7/69.6 & 68.0/66.9 \\
			Point2Seq\cite{point2seq} & One        & -      & -/-     & -/-     & 77.5/77.0 & 68.8/68.4 & -/-     & -/-     & -/-     & -/- \\
			CenterFormer\cite{centerformer} & One        &  -     & 75.3/72.9 & 71.1/68.9 & 75.0/74.4 & 69.9/69.4 & 78.6/73.0 & 73.6/68.3 & 72.3/71.3 & 69.8/68.8 \\
			PillarNet-34\cite{pillarnet} & One        & -      & 77.3/74.6 & 71.0/68.5 & 79.1/78.6 & 70.9/70.5 & 80.6/74.0 & 72.3/66.2 & 72.3/71.2 & 69.7/68.7 \\
			SWFormer\cite{swformer} & One       & -      & -/-     & -/-     & 77.8/77.3 & 69.2/68.8 & 80.9/72.7 & 72.5/64.9 & -/-     & -/- \\
			AFDetV2\cite{afdetv2} & One        &  -     & 77.2/74.8 & 71.0/68.8 & 77.6/77.1 & 69.7/69.2 & 80.2/74.6 & 72.2/67.0 & 73.7/72.7 & 71.0/70.1 \\
			ConQueR\cite{conquer} & One       & 14.3  & 76.3/73.5 & 70.3/67.7 & 76.1/75.6 & 68.7/68.2 & 79.0/72.3 & 70.9/64.7 & 73.9/72.5 & 71.4/70.1 \\
			GD-MAE\cite{gdmae} & One        &  -     & 76.9/73.7 & 70.6/67.6 & 77.3/76.7 & 68.7/68.3 & 80.3/72.4 & 72.8/65.5 & 73.1/71.9 & 70.3/69.2 \\
			VoxelNeXt\cite{voxelnext} & One        &  -     & 78.6/76.3 & 72.2/70.1 & 78.2/77.7 & 69.9/69.4 & 81.5/76.3 & 73.5/68.6 & 76.1/74.9 & 73.3/72.2 \\
			DSVT-P\cite{dsvt} & One        & 14.9  & 79.5/77.1 & 73.2/71.0 & 79.3/78.8 & 70.9/70.5 & 82.8/77.0 & 75.2/69.8 & 76.4/75.4 & 73.6/72.7 \\
			DSVT-V\cite{dsvt} & One        & 10.3  & 80.3/78.2 & 74.0/72.1 & 79.7/79.3 & 71.4/71.0 & 83.7/78.9 & 76.1/71.5 & 77.5/76.5 & 74.6/73.7 \\
			\hline
			Part-A2-Net\cite{parta2} & Two        & -      & 73.6/70.3 & 66.9/63.8 & 77.1/76.5 & 68.5/68.0 & 75.2/66.9 & 66.2/58.6 & 68.6/67.4 & 66.1/64.9 \\
			PV-RCNN\cite{pvrcnn} & Two        & -      & 76.2/73.6 & 69.6/67.2 & 78.0/77.5 & 69.4/69.0 & 79.2/73.0 & 70.4/64.7 & 71.5/70.3 & 69.0/67.8 \\
			LiDAR-RCNN\cite{lidarrcnn} & Two        & -      & 71.9/63.9 & 65.8/61.3 & 76.0/75.5 & 68.3/67.9 & 71.2/58.7 & 63.1/51.7 & 68.6/66.9 & 66.1/64.4 \\
			PointAugmenting\cite{pointaugmenting} & Two    & -      & 72.9/ - & 66.7/ - & 67.4/ - & 62.7/ - & 75.0/ - & 70.6/ - & 76.3/ - & 74.4/ - \\
			SST-TS\cite{sst} & Two        & -      & -/-     & -/-     & 76.2/75.8 & 68.0/67.6 & 81.4/74.0 & 72.8/65.9 & -     & - \\
			PDV\cite{pdv}   & Two        &  -     & 73.3/70.0 & 67.2/64.2 & 76.9/76.3 & 69.3/68.8 & 74.2/66.0 & 65.9/58.3 & 68.7/67.6 & 66.5/65.4 \\
			PV-RCNN++$\ast$\cite{pvrcnn++} & Two        & 10    & 78.1/75.9 & 71.7/69.5 & 79.3/78.8 & 70.6/70.2 & 81.3/76.3 & 73.2/68.0 & 73.7/72.7 & 71.2/70.2 \\
			FSD\cite{fsd}   & Two        & -      & 79.6/77.4 & 72.9/70.8 & 79.2/78.8 & 70.5/70.1 & 82.6/77.3 & 73.9/69.1 & 77.1/76.0 & 74.4/73.3 \\
			OcTr\cite{octr}  & Two        & -      & 77.2/74.5 & 70.7/68.2 & 78.1/77.6 & 69.8/69.3 & 80.8/74.4 & 72.5/66.5 & 72.6/71.5 & 69.9/68.9 \\
			LoGoNet\cite{logonet} & Two    & -      & 79.5/77.0 & 73.7/71.4 & 79.0/78.4 & 71.2/70.7 & 82.9/77.1 & 75.5/69.9 & 76.6/75.5 & 74.5/73.5 \\
			DSVT-V-TS\cite{dsvt} & Two        & -      & 81.1/78.9 & 74.8/72.8 & 80.4/79.9 & 72.2/71.8 & \textbf{84.2/79.3} & 76.5/\textbf{71.8} & 78.6/77.6 & 75.7/74.7 \\
			\hline
			DIRM (Ours) & One            &34.6 & 79.8/77.3 & 73.7/71.3 & 78.9/78.4 & 71.1/70.6 & 82.9/77.0 & 75.2/69.6 & 77.7/76.5 & 74.9/73.8 \\
			DIRM$\ddagger$ (Ours) & One           &19.5 & \textbf{81.4/78.9} & \textbf{75.4/73.1} & \textbf{80.5/80.0} & \textbf{72.4/71.9}  & 83.7/78.0 & \textbf{76.6}/71.1 & \textbf{79.9/78.8} & \textbf{77.4/76.3} \\
			\hline
	    \end{tabular}}
         \caption{3D object detection performance of DIRM compared to state-of-the-art methods on the Waymo \textit{val} set. $\ast$ denotes PV-RCNN++ with center head and $\ddagger$ represents DIRM with a deeper backbone. The baseline method is CenterPoint and \textbf{bolded} values denote the highest mAP/mAPH.}
         \vspace{-0.35cm} 
    	\label{tab1}%
\end{table*}%

\begin{figure}[h!]
    \centering
    \includegraphics[width=\columnwidth]{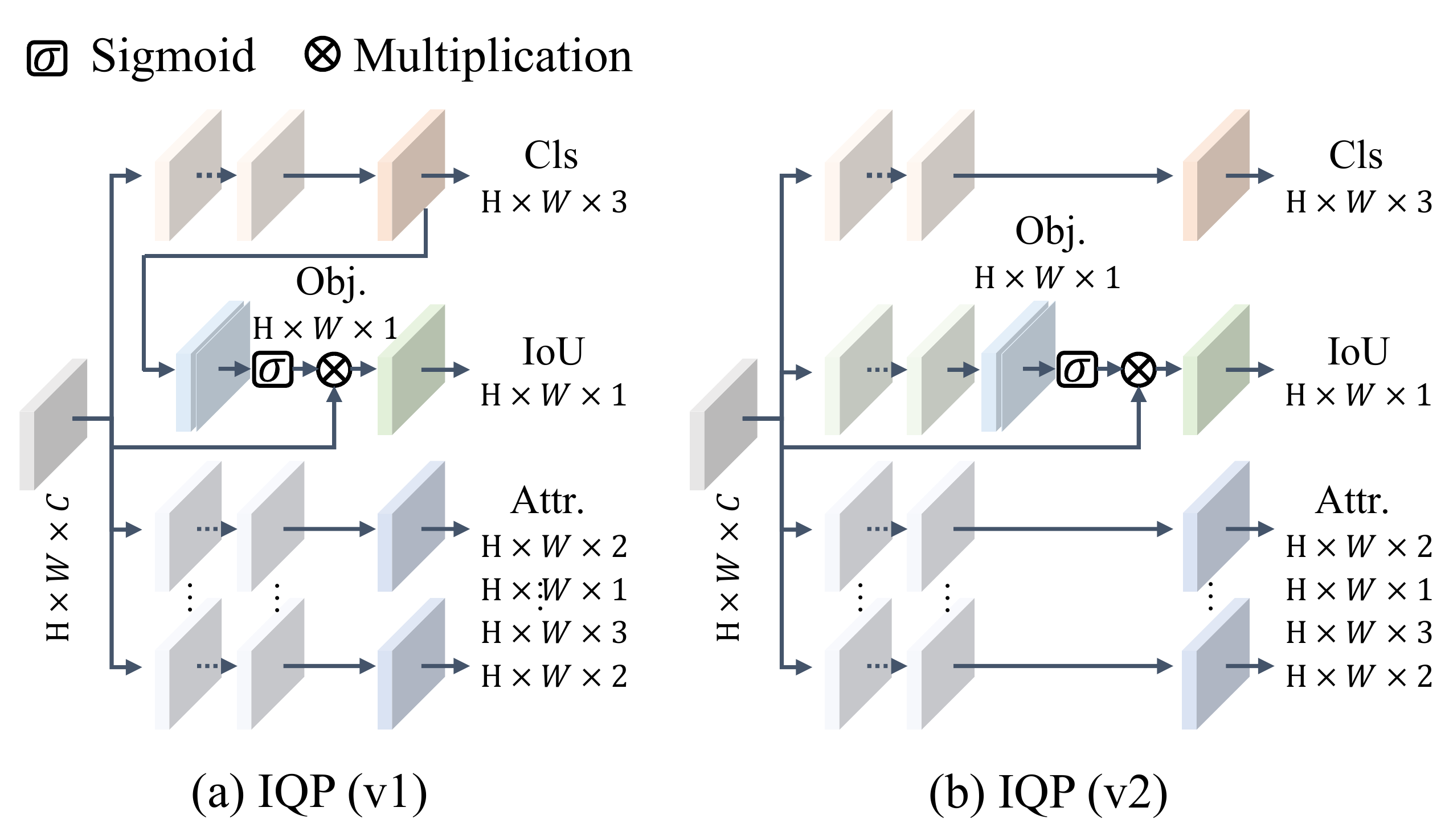}
    \caption{Schematic diagram of interactive quality prediction. "Cls", "Attr.", and "Obj." denote the classification branch, bounding box attribute regression branch, and class-agnostic object classification branch, respectively.}
    \vspace{-0.35cm} 
    \label{fig5}
\end{figure}

\subsection{Interactive Quality Prediction (IQP)}

Inconsistency between classification scores and regression quality is a common issue in detection tasks. In 3D object detection, rectifying the classification score with the quality (IoU) prediction branch has been proven to be an effective solution. However, as discussed earlier, existing IoU regression modeling approaches provide unreliable IoU predictions for low-quality prediction results, leading to suboptimal performance. In fact, these low-quality predictions are typically defined as negative samples, obtaining sufficient supervision in classification tasks. Inspired by this, IQP attempts to introduce the classification task to interact with the quality regression task for joint optimization.

{\em \textbf{Class-agnostic Classification}}. To facilitate interaction with the quality prediction information $F_{IoU}\in\mathbb{R}^{[ 1\times H\times W]}$, IQP first constructs class-agnostic classification information $F_{obj}\in\mathbb{R}^{[1\times H\times W]}$ (foreground and background classification information).
To obtain $F_{obj}$, we propose two strategies. A simple strategy, IQP (v1), involves compressing the category channels of the original classification branch $F_{conf}\in\mathbb{R}^{[C\times H\times W]}$ as follows (Fig. \ref{fig5} (a)):
\begin{equation}
\label{equation_3}
F_{obj}=Max(F_{conf}, dim=1),
\end{equation}
where $Max$ denotes the maximum function, and $C$ is the number of categories. The probability that a location is an object can be approximated by calculating the maximum value. The other strategy, IQP (v2), involves producing $F_{obj}$ via an independent branch as follows (Fig. \ref{fig5} (b)):
\begin{equation}
\label{equation_4}
F_{obj}=Conv_1(Conv_2(X)),X\in \mathbb{R}^{[64 \times H \times W]},
\end{equation}
where $X$ denotes the shared BEV features, and $Conv$ is the convolution operation. As shown in Fig. \ref{fig3}, $F_{obj}$ can also generate specific feature sizes using two convolutions, similar to other branches in the detection head. The supervision signal for $F_{obj}$ is the compression of the original classification label $Label_{conf} \in \mathbb{R}^{[C\times H\times W]}$ as same as Equation \ref{equation_3}.

Binary Cross-Entropy (BCE) loss is used to calculate the loss values of the class-agnostic classification. The effects of the two strategies are compared in the \textit{Ablation Studies}.

{\em \textbf{Tasks Interaction}}. With the class-agnostic classification task modeled as described above, $F_{obj}$ possesses the basic ability to classify the foreground and background. The background classification ability, lacking in the regression task, can be utilized to suppress the erratic output of those low-quality predictions in the quality prediction task. As shown in Fig. \ref{fig5} (a, b), IQP interacts class-agnostic classification tasks with regression tasks through a cascading approach:
\begin{equation}
\label{equation_5}
F_{IoU}=Conv_3(Sigmoid(F_{obj}) \times X).
\end{equation}

The quality supervision signal comes from the prediction box of the positive sample and the IoU value $IoU_{GT}$ of GT:
\begin{equation}
\label{equation_6}
Label_{IoU}=2 \times IoU_{GT} - 1, IoU_{GT} \in \mathbb{R}^{N_{pos}},
\end{equation}
where $N_{pos}$ denotes the number of positive samples. Following PillarNet\cite{pillarnet} and SA-SSD\cite{sassd}, IQP adopts $L1 \ loss$ to calculate the loss value for sparse positive samples.

IQP ensures reliable IoU predictions for prediction results of varying quality by jointly optimizing object classification and quality regression tasks. In contrast to directly regressing IoU, IQP tends to improve the IoU prediction accuracy of low-quality prediction results. Besides, IQP is highly succinct and efficient, introducing only two additional convolution operations that have minimal impact on the inference speed.

\section{Experiments and Results}
\label{sec4}

\subsection{Datasets and Evaluation}

For comparisons with SOTA methods, results are presented trained on 100\% of the Waymo \cite{waymo} and ONCE \cite{once} \textit{val} sets. Ablation and generalization experiments are conducted on the Waymo dataset using 20\% of the training set. Official evaluation metrics for both datasets are applied to evaluate method performance.

\subsection{Implementation and Details}

{\em \textbf{Data Preprocessing}}. To accommodate irregular point clouds into the detector, the original point cloud space is voxelized. For Waymo/ONCE datasets, detection ranges and voxel sizes are set as [-75.2, 75.2], [-75.2, 75.2], [-2.0/-5.0, 4.0/3.0] m and (0.1, 0.1, 0.15/0.2) m, respectively. DIRM uses common data augmentation strategies \cite{centerpoint,dsvt}.

{\em \textbf{Training Details}}. DIRM employs the Adam optimizer for end-to-end optimization, adopting the One-cycle strategy. The division coefficient is 10, the momentum range is [0.95, 0.85], and the weight decay rate is 0.05. The maximum learning rate on both datasets is set to 0.003. DIRM is trained on the Waymo dataset for 30 epochs. As for the ONCE dataset, the detector is trained for 80 epochs. Experiments are conducted on 4 NVIDIA A100 GPUs with 40 GB memory, and the batch size is set to 16 for both datasets.

\begin{table*}[t]
	\centering
	\resizebox{0.9\linewidth}{!}{
		\begin{tabular}{l llll llll llll l}
			\hline
			\multirow{2}{*}{Methods} & \multicolumn{4}{c}{Vehicle} & \multicolumn{4}{c}{Pedestrian} & \multicolumn{4}{c}{Cyclist} & \multirow{2}{*}{mAP} \\ 
			& Overall & 0$\textendash$30 m & 30$\textendash$50 m & $>$50 m & Overall & 0$\textendash$30 m & 30$\textendash$50 m & $>$50 m & Overall & 0$\textendash$30 m & 30$\textendash$50 m & $>$50 m   & \\ \hline
			PointRCNN$\dagger$\cite{pointrcnn} & 52.09 & 74.45 & 40.89 & 16.81 & 4.28  & 6.17  & 2.40   & 0.91  & 29.84 & 46.03 & 20.94 & 5.46 & 28.74 \\
			PointPillars$\dagger$\cite{pointpillars} & 68.57 & 80.86 & 62.07 & 47.04 & 17.63 & 19.74 & 15.15 & 10.23 & 46.81 & 58.33 & 40.32 & 25.86 & 44.34 \\
			SECOND$\dagger$\cite{second} & 71.19 & 84.04 & 63.02 & 47.25 & 26.44 & 29.33 & 24.05 & 18.05 & 58.04 & 69.96 & 52.43 & 34.61 & 51.89 \\
			PV-RCNN$\dagger$\cite{pvrcnn}  & 77.77 & 89.39 & 72.55 & 58.64 & 23.50  & 25.61 & 22.84 & 17.27 & 59.37 & 71.66 & 52.58 & 36.17 & 53.55\\
			BSAODet\cite{bsaodet}  & 78.81 & \underline{89.47} & 72.88 & 58.64 & 27.72 & 32.41 & 24.54 & 16.40  & 60.60  & 73.86 & 53.36 & 36.98 & 55.71\\
			IA-SSD\cite{iassd} & 70.30  & 83.01 & 62.84 & 47.01 & 39.82 & 47.45 & 32.75 & 18.99 & 62.17 & 73.78 & 56.31 & 39.53 & 57.43 \\
			DBQ-SSD\cite{dbqssd} & 72.14 & 84.81 & 64.27 & 50.22 & 37.83 & 43.88 & 32.18 & 20.29 & 62.99 & 75.13 & 56.65 & 38.91 & 57.65 \\
			PointPainting$\dagger$\cite{pointpainting}  & 66.17 & 80.31 & 59.80  & 42.26 & 44.84 & 52.63 & 36.63 & 22.47 & 62.34 & 73.55 & 57.20  & 40.39 & 57.78\\
			IC-FPS\cite{icfps} & 70.56 & 82.73 & 64.47 & 48.75 & 40.09 & 47.64 & 32.57 & 20.51 & 62.80  & 75.64 & 57.65 & 38.14 & 57.82 \\
			CenterPoint$\dagger$\cite{centerpoint} & 66.79 & 80.10  & 59.55 & 43.39 & 49.90  & 56.24 & 42.61 & 26.27 & 63.45 & 74.28 & 57.94 & 41.48 & 60.05 \\
			CG-SSD\cite{cgssd} & 67.6  & 80.22 & 61.23 & 44.77 & 51.50  & 58.72 & 43.36 & 27.76 & 65.79 & 76.27 & 60.84 & 43.35 & 61.63 \\
			Point2Seq\cite{point2seq} & 73.43 & 85.16 & 66.21 & 50.76 & 57.53 & 68.21 & 47.15 & 25.18 & 67.53 & 77.95 & 62.14 & 46.06 & 66.16 \\
			GD-MAE\cite{gdmae} & 75.64 & 87.21 & 70.10  & 53.21 & 45.92 & 54.78 & 37.84 & 22.56 & 66.30  & 78.12 & 60.52 & 42.05 & 62.62 \\
			CenterPoint $\ast$\cite{centerpoint} & 76.33 & 86.39 & 71.74 & 59.98 & 51.35 & 60.27 & 43.80  & 25.21 & 67.98 & 79.17 & 62.38 & 46.02 & 65.22 \\
			\hline
			DIRM (Ours)     & \underline{79.99} & 89.06 & \underline{75.55} & \underline{63.09} & \underline{59.16} & \underline{69.50}  & \underline{49.25} & \underline{28.69} & \underline{69.82} & \underline{80.33} & \underline{64.72} & \underline{48.12} & \underline{69.66} \\
			DIRM$\ddagger$ (Ours)     & \textbf{81.95} & \textbf{90.80}  & \textbf{77.51} & \textbf{65.23} & \textbf{59.56} & \textbf{70.19} & \textbf{49.65} & \textbf{28.25} & \textbf{71.75} & \textbf{82.15} & \textbf{66.42} & \textbf{51.16} & \textbf{71.09} \\
			\hline
  \end{tabular}}%
   \caption{3D object detection performance of DIRM compared to state-of-the-art methods on the ONCE \textit{val} set. 
    $\dagger$ denotes the official performance benchmark provided by the ONCE. 
    $\ast$ represents detection performance reproduced using a publicly released code. $\ddagger$ denotes DIRM with a deeper backbone and \textbf{bolded} values denote the highest mAP/mAPH. \underline{Underlined} values denote the second-highest performance. The baseline method is CenterPoint.}
    \vspace{-0.35cm} 
	\label{tab2}%
\end{table*}%

\subsection{Comparison with State-of-the-Art Detecotrs}

Tab. \ref{tab1} exhibits the performance of DIRM compared with current SOTA methods on the Waymo \textit{val} set. DIRM achieves new SOTA performance in terms of both inference speed and accuracy, outperforming methods with the same backbone network. For instance, DIRM leads the latest one-stage method ConQueR \cite{conquer} and prevailing PV-RCNN++ \cite{pvrcnn++} by 3.44/3.61 and 4.14/4.11 L2 mAP/mAPH, respectively. For the baseline method \cite{centerpoint}, DIRM improves the performance by 5.54/5.51 L2 mAP/mAPH with little additional time consumption (34.6 vs. 34.8 FPS). 

Notably, the proposed DIRM$\ddagger$ with a deeper backbone outperforms all current one-stage methods, including the SOTA method DSVT \cite{dsvt} which uses a transformer backbone by 1.44/1.01 L2 mAP/mAPH. Besides, the inference speed of DIRM$\ddagger$ is twice that of DSVT (19.5 vs. 10.3). DIRM$\ddagger$ respectively leads the SOTA multi-modal method LoGoNet \cite{logonet} and the two-stage method DSVT-V-TS \cite{dsvt} by 1.74/1.71 and 0.64/0.31 L2 mAP/mAPH. Owing to the significant improvement in localization accuracy achieved by DAR and IQP, DIRM$\ddagger$ attains the SOTA performance in the vehicle and cyclist categories with larger sizes.

For the ONCE \textit{val} set, DIRM significantly improves the baseline \cite{centerpoint} by 4.44 mAP, particularly for distant objects with sparse point clouds, as shown in Tab. \ref{tab2}. For instance, DIRM respectively outperforms the baseline by 3.81, 5.45, and 2.34 AP on the vehicle, pedestrian, and cyclist categories, respectively, in the range of 30-50 meters. Besides, DIRM and DIRM$\ddagger$ significantly outperform the previous SOTA method Point2Seq \cite{point2seq} by more than 3.5 mAP, achieving the best performance.

\subsection{Ablation Studies}

The effect of each DIRM component on the Waymo \textit{val} set is exhibited in Tab \ref{tab3}. CenterPoint \cite{centerpoint} is employed as the baseline method. As shown in the second and third rows, DAR and IQP boost the baseline by 2.08/1.85 and 3.57/3.59 mAP/mAPH, respectively. Combining DAR and IQP can further enhance the baseline performance by 4.98/4.91 mAP/mAPH. The above ablation experiments reveal that DAR and IQP are crucial for performance improvement in all categories. Detailed ablation experiments are conducted on each module by systematically peeling from top to bottom, as discussed in the sequel and the supplemental.

\begin{table}[t]
	\centering
	\resizebox{0.95\linewidth}{!}{
		\begin{tabular}{l cc c ccc}
			\hline
			& \multirow{2}{*}{DAR} & \multirow{2}{*}{IQP} & \multirow{2}{*}{L2 mAP/mAPH} & Vehicle & Pedestrian & Cyclist \\
			& & & & L2 AP/APH & L2 AP/APH & L2 AP/APH \\
			\hline
			(a) & & & 66.47/64.02 & 64.88/64.37 & 66.54/60.92 & 67.99/66.78 \\
			(b) & \checkmark& & 68.55/65.87 & 66.81/66.29 & 68.28/62.00 & 70.56/69.32 \\
			(c) & & \checkmark & 70.04/67.61 & 67.3/66.85 & 71.05/65.34 & 71.77/70.65 \\
			(d) & \checkmark & \checkmark& \textbf{71.45/68.93} & \textbf{69.4/68.90} & \textbf{72.28/66.39} & \textbf{72.68/71.50} \\
			\hline
    \end{tabular}}%
    \caption{Effect of components in the proposed DIRM.}
    \vspace{-0.35cm} 
	\label{tab3}%
\end{table}%

{\em \textbf{DAR}}. To showcase the effectiveness of DAR, experiments involving four types of sample-assignment strategies are conducted (Tab. \ref{tab4}). As shown in the first row, directly increasing the number of samples cannot improve the performance and even seriously damages the performance of the pedestrian and cyclist categories. It can be inferred from the second and third rows that the dynamic assignment outperforms the static one by 0.75/0.74 AP/APH on the best-benefited category (vehicle). This phenomenon demonstrates that modeling with adaptively optimal sample selection can effectively alleviate the burden of the multi-sample strategy when calculating the loss. The comparison between the third and fourth rows demonstrates that DAR (switch) is an advanced strategy (Fig. \ref{fig4}). This is because in the early training stages, quality predictions are not reliable, and the static sample allocation strategy can ensure the stability of early training. Thus, DAR (switch) is the final solution.

\begin{table}[h!]
    \vspace{-0.15cm} 
	\centering
	\resizebox{0.95\linewidth}{!}{
    \begin{tabular}{l c ccc}
        \hline
        \multirow{2}{*}{Method} & \multirow{2}{*}{L2 mAP/mAPH} & Vehicle & Pedestrian & Cyclist \\
        & & L2 AP/APH & L2 AP/APH & L2 AP/APH \\
        \hline
        Multi-pos & 66.41/64.04 & 68.48/68.04 & 62.37/56.84 & 68.39/67.24 \\
        DAR (static) & 70.99/68.47 & 68.62/68.14 & 72.27/66.37 & 72.07/70.91 \\
        DAR (dynamic) & 71.26/68.71 & 69.37/68.88 & \textbf{72.30}/66.29 & 72.11/70.95 \\
        DAR (switch) & \textbf{71.45/68.93} & \textbf{69.40/68.90} & 72.28/\textbf{66.39} & \textbf{72.68/71.50} \\
        \hline
    \end{tabular}}%
    \caption{Comparison of different sample assignment strategies.}
    \vspace{-0.25cm} 
	\label{tab4}%
\end{table}

{\em \textbf{IQP}}. To illustrate the effectiveness of IQP, experiments are conducted using different quality interaction strategies. As shown in Tab. \ref{tab5}, the comparison between the first and second rows demonstrates that interacting with class-agnostic classification information can effectively improve the regression modeling quality, especially for the pedestrian category (+3.60/4.10 L2 mAP/mAPH). This is because using unreliable quality predictions to rectify the confidence scores can cause high response locations to deviate from the true object, leading to errors of a few decimeters, and greatly limiting the performance of small-sized categories. The results of the second and third rows indicate that establishing an independent class-agnostic classification prediction branch can provide more accurate object confidence and significantly improve detection performance (+0.36/0.31 L2 mAP/mAPH). Thus, we choose IQP (v2) as the final solution.

\begin{table}[h!]
    \vspace{-0.15cm} 
	\centering
	\resizebox{0.95\linewidth}{!}{
	\begin{tabular}{l c ccc}
		\hline
		\multirow{2}{*}{Method} & \multirow{2}{*}{L2 mAP/mAPH} & Vehicle & Pedestrian & Cyclist \\
		& & L2 AP/APH & L2 AP/APH & L2 AP/APH \\
		\hline
		w/o IQP & 68.55/65.87 & 66.81/66.29 & 68.28/62.00 & 70.56/69.32 \\
		IQP (v1) & 71.09/68.62 & 69.03/68.55 & 71.88/66.10 & 72.35/71.20 \\
		IQP (v2) & \textbf{71.45/68.93} & \textbf{69.40/68.90} & \textbf{72.28/66.39} & \textbf{72.68/71.50} \\
		\hline
	\end{tabular}}%
    \caption{Comparison of different quality interaction strategies}
    \vspace{-0.45cm} 
	\label{tab5}%
\end{table}%

\subsection{Generalization Capacity}

To assess the generalization performance of DIRM, we extend it to mainstream methods with varying point cloud representations and detection stages. As shown in Tab. \ref{tab6}, DIRM brings substantial improvements to pillar-based methods such as CenterPoint (Pillar) \cite{centerpoint} and PillarNet \cite{pillarnet}, enhancing them by 3.63/3.66 L2 mAP/mAPH and 1.98/1.96 L2 mAP/mAPH, respectively. Notably, for pillar-based methods, DIRM exhibits particularly significant performance gains in the vehicle category, enhancing CenterPoint (Pillar) 4.63/4.60 L2 AP/APH. It reveals that inaccurate localization has a greater impact on the vehicle category for pillar-based methods. In addition, DIRM extends its effectiveness beyond one-stage methods, significantly boosting the previous SOTA two-stage method PV-RCNN++ \cite{pvrcnn++} by 1.84/1.88 L2 mAP/mAPH.

These findings highlight the remarkable generalization ability of DIRM, demonstrating its applicability to center-based methods with diverse point cloud representation and detection stages. Please refer to the supplemental for more experimental results on different datasets.

\begin{table}[h!]
	\centering
	\resizebox{0.95\linewidth}{!}{
	\begin{tabular}{l c ccc}
		\hline
		\multirow{2}{*}{Method} & \multirow{2}{*}{L2 mAP/mAPH} & Vehicle & Pedestrian & Cyclist \\
		& & L2 AP/APH & L2 AP/APH & L2 AP/APH \\
		\hline
		CenterPoint & 66.47/64.02 & 64.88/64.37 & 66.54/60.92 & 67.99/66.78 \\
		W/ Ours & 71.45/68.93 & 69.40/68.90 & 72.28/66.39 & 72.68/71.50 \\
		Improvement & \textbf{+4.98/+4.91}&  \textbf{+4.52/+4.53}&  \textbf{+5.74/+5.47}&  \textbf{+4.69/+4.72}  \\
		\hline
		CenterPoint (Pillar) & 63.90/60.05 & 62.06/61.58 & 65.91/56.33 & 63.73/62.24 \\
		W/ Ours & 67.53/63.71 & 66.69/66.18 & 70.46/60.91 & 65.44/64.03 \\
		Improvement & \textbf{+3.63/+3.66}&  \textbf{+4.63/+4.6}& \textbf{+4.55/+4.58}&  \textbf{+1.71/+1.79}  \\
		\hline
		PillarNet & 67.85/64.78 & 66.86/66.38 & 69.76/62.27 & 66.92/65.69 \\
		W/ Ours & 69.83/66.74 & 69.14/68.66 & 70.71/63.14 & 69.64/68.41 \\
		Improvement & \textbf{+1.98/+1.96}&  \textbf{+2.28/+2.28}&  \textbf{+0.95/+0.87}&  \textbf{+2.72/+2.72}  \\
		\hline
		PV-RCNN++ & 69.86/67.35 & 69.02/68.56 & 71.42/65.44 & 69.14/68.06 \\
		W/ Ours & 71.70/69.23 & 69.96/69.51 & 73.17/67.30 & 71.97/70.88 \\
		Improvement & \textbf{+1.84/+1.88} & \textbf{+0.94/+0.95}&  \textbf{+1.75/+1.86}&  \textbf{+2.83/+2.82}  \\
		\hline
	\end{tabular}}%
    \caption{Performance of extending DIRM to mainstream methods on the Waymo \textit{val} Set}
    \vspace{-0.25cm} 
	\label{tab6}%
\end{table}%

\subsection{Quantitative Analysis}

\begin{figure}[t]
    \centering
    \includegraphics[width=\columnwidth]{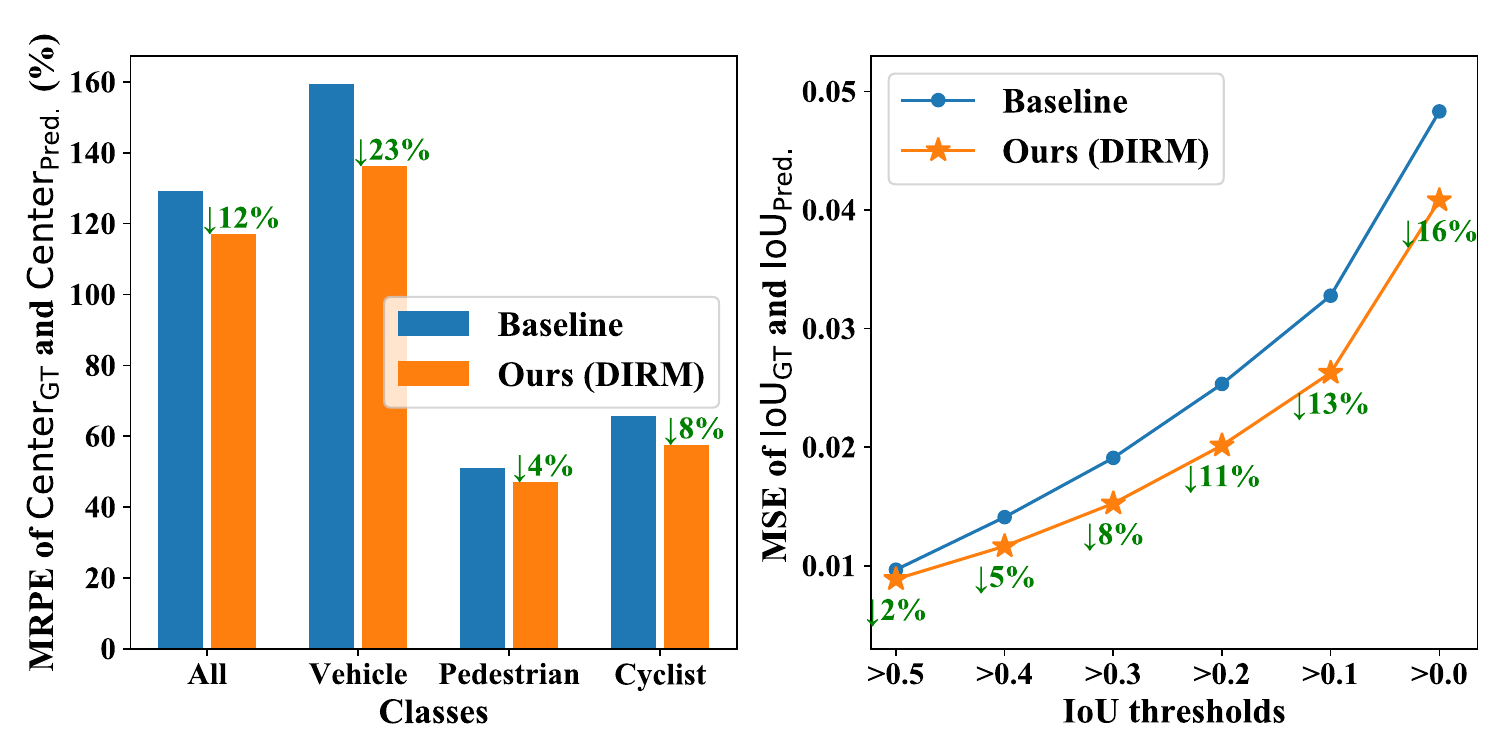}
    \caption{Quality prediction and quantitative analysis of center attribute regression.}
    \vspace{-0.35cm} 
    \label{fig6}
\end{figure}

To further validate DIRM’s ability to rectify the inaccurate regression of the center attribute, we examine the MRPE of the center attribute. As illustrated in Fig. \ref{fig6} (a), the baseline method \cite{centerpoint} exhibits an average relative error of 130\% for all categories. DIRM significantly reduces this error by 12\%. Notably, for the vehicle category, the baseline method reaches an error of 160\%, and DIRM substantially mitigates it by 23\%. 
In addition, we assess the MSE of the predicted quality compared to the real quality under different thresholds to investigate the impact of DIRM on the accuracy of quality prediction. As illustrated in Fig. \ref{fig6} (b), the improvement of DIMR on quality prediction becomes more pronounced as the quality threshold decreases. 
DIRM reduces the MSE by 16\% compared to the baseline when the threshold is zero. In summary, DIRM not only significantly enhances performance, but also effectively improves the regression accuracy of the center attribute and quality.

Besides, DIRM maintains consistent inference latency, model parameters, and floating-point operands with the baseline, while achieving a substantial improvement of 4.98/4.91 L2 mAP/mAPH (Tab. \ref{tab7}). In comparison to the SOTA transformer-based method \cite{dsvt}, DIMR$\ddagger$ exhibits nearly half the inference latency of DSVT-V \cite{dsvt}, fewer than 2/5 of the floating-point operands, and a noteworthy performance improvement of 1.99/1.88 L2 mAP/mAPH.

These results demonstrate the outstanding overall performance of DIRM in terms of inference latency, model parameters, and detection performance. 

\begin{table}[h!]
	\centering
	\resizebox{0.75\linewidth}{!}{
	\begin{tabular}{l c c c c c}
		\hline
		\multirow{2}{*}{Method} & Latency & Parmas & FLOPs & L2  \\
		& (ms) & (MB) & (GB) & mAP/mAPH\\
		\hline
		CenterPoint & 28.76 & 5.07  & 93.27 &  66.47/64.02 \\
		PV-RCNN++ & 100.09 & 13.41 & 101.80 &  69.86/67.35 \\
		DSVT-P & 67.00    & 7.47  & 519.63 &  71.14/68.59 \\
		DSVT-V & 97.00    & 7.47  & 522.82 &  72.01/69.67 \\
		\hline
		DIRM (Ours) & 28.87 & 5.14  & 95.93 &  71.45/68.93 \\
		DIRM$\ddagger$ (Ours) & 51.39 & 6.35  & 208.90 &  74.00/71.55 \\
		\hline
	\end{tabular}}%
    \caption{Quantification analysis of DIRM and state-of-the-art methods}
	\label{tab7}%
    \vspace{-0.35cm} 
\end{table}%

\section{Conclusion}

This study indicates that the primary issue hindering the performance of center-based one-stage detectors is the incomplete modeling of the center attribute and the quality regression task. To fully unleash the potential of one-stage pipelines based on the above observation, this study introduces two novel components, Decoupled Attribute Regression (DAR) and Interactive Quality Prediction (IQP). Specifically, DAR establishes long-range regression modeling for the center attribute through deep decoupling of bounding box attributes and an independent adaptive multi-sample assignment strategy. On the other hand, IQP optimizes quality predictions by incorporating object classification information, which is proficient in modeling negative samples, to furnish reliable IoU predictions for low-quality predictions. With the plug-and-play DAR and IQP components, we propose a high-performance one-stage detection framework, DIRM, that is comparable to two-stage methods and can be easily integrated into any center-based method. Extensive experiments on the Waymo and ONCE datasets demonstrate that DIRM achieves outstanding comprehensive performance concerning inference latency, model parameters, detection performance, and generalization performance. In particular, DIRM obtains SOTA one-stage detection performance on both datasets, surpassing previous SOTA two-stage methods.

\bibliographystyle{named}
\bibliography{ijcai24}

\end{document}